\documentclass{article}

     \PassOptionsToPackage{numbers, compress}{natbib}

\usepackage[final]{neurips_2019}

\usepackage[utf8]{inputenc} 
\usepackage[T1]{fontenc}    
\usepackage{hyperref}       
\usepackage{url}            
\usepackage{booktabs}       
\usepackage{amsfonts}       
\usepackage{nicefrac}       
\usepackage{microtype}      
\usepackage{amsmath}
\usepackage{graphicx}
\usepackage{hhline}
\usepackage{multirow}
\usepackage{algorithm}
\usepackage[noend]{algpseudocode}
\usepackage{xspace}

\usepackage{subfig}

\newcommand\tab[1][0.8cm]{\hspace*{#1}}

\newcommand{\prjname}{MUTE\xspace}

\title{\prjname: Data-Similarity Driven \underline{Mu}lti-hot \underline{T}arget \underline{E}ncoding for Neural Network Design}

\author{
    Mayoore S. Jaiswal\textsuperscript{1} \tab 
    Bumsoo Kang\textsuperscript{2}\thanks{Word done during internship at IBM} \tab
    Jinho Lee\textsuperscript{3} \thanks{Work done while at IBM} \tab 
    Minsik Cho\textsuperscript{1} \\
    \textsuperscript{1}IBM, Austin, TX, USA \tab 
    \textsuperscript{2}KAIST, Daejeon, South Korea \tab 
    \textsuperscript{3}Yonsei University, Seoul, South Korea \\
    \texttt{mayoore.s.jaiswal@ibm.com, minsikcho@us.ibm.com} \\
}

\begin{document}

\maketitle
\begin{abstract}
Target encoding is an effective technique to deliver better performance for conventional machine learning methods, and recently, for deep neural networks as well. However, the existing target encoding approaches require significant increase in the learning capacity, thus demand higher computation power and more training data.
In this paper, we present a novel and efficient target encoding scheme, \prjname to improve both generalizability and robustness of a target model by understanding the inter-class characteristics of a target dataset. By extracting the confusion level between the target classes in a dataset, \prjname strategically optimizes the Hamming distances among target encoding. Such optimized target encoding offers higher classification strength for neural network models with negligible computation overhead and without increasing the model size.
When \prjname is applied to the popular image classification networks and datasets, our experimental results show that \prjname offers better generalization and defense against the noises and adversarial attacks over the existing solutions. 
\end{abstract}

\section{Introduction}
\label{sec:intro}

Scalable artificial intelligent systems require a methodology for efficient neural network design that can generalize well, learn semantics of the training dataset, and resist adversarial attacks. However, existing methods have been shown to learn dataset bias~\cite{torralba2011unbiased, tommasi2017deeper, li2014secrets}, and failed to deliver sufficient generalization capability. Poor generalization makes models unpredictable, causes potential ethical issues, and mis-guides neural network design~\cite{rethinkgeneralization,datasheet4dataset}. To tackle the generalization problems, target encoding has been studied for both conventional machine learning and deep neural network architectures and proven to be highly effective~\cite{akata2016label,frome2013devise, hsu2009multi, cisse2013robust}. Yet, many prior works in target encoding require a long encoding sequence (which increases the model size) and fail to tailor the encoding for a given task or dataset. Furthermore, they do not investigate the effects of different target encodings against noisy data and adversarial attacks.

\begin{figure}[t]
	\centering
	\includegraphics[width=0.8\textwidth]{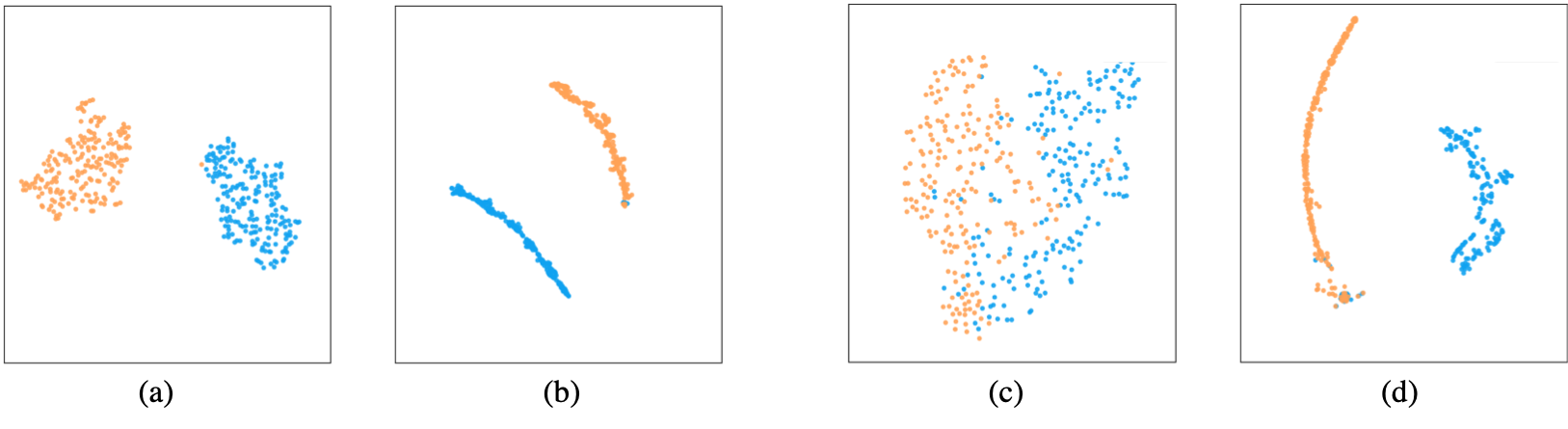}
	\caption{TSNE~\cite{maaten2008visualizing} visualizations of a ConvNet (refer Section~\ref{sec:mute_exp} for architecture) with one-hot encoding trained on MNIST (a, c) and a ConvNet with \prjname trained on MNIST (b, d). Classes well-separated in features space in models trained with one-hot encodings (a), remain well separated in models trained with \prjname (b). However, class not-well-separated in features space in models trained with one-hot encodings (c), are well-separated in feature space when trained with \prjname.}
	\label{fig:overview}
\end{figure}

In this work, we propose \prjname, a systematic approach to make deep learning models generalize better by optimizing the target encoding~\cite{akata2016label,frome2013devise, hsu2009multi, cisse2013robust}. 
Unlike the conventional one-hot method where the Hamming distance between labels is fixed at $2$, \prjname generates a multi-hot encoding by exploiting the expression power of a given output encoding length.
\prjname strategically extracts the \textit{similarity} between pairs of classes from a dataset, and leverages that information to
obtain a multi-hot encoding such that semantically closer classes are forced to be further apart in the label space in terms of the Hamming distance. \prjname ensures that Stochastic Gradient Decent (SGD) algorithm extracts distinctive features between two easy-to-confuse classes, which in turn reduces the chance of mis-prediction under noisy and noiseless conditions.
Figure~\ref{fig:overview} illustrates the high-level idea in \prjname, that is it increases the distance between classes in feature space.
In order to find such high-quality encoding in \prjname, we formulate the encoding generation as a subgraph isomorphism problem~\cite{ullman1976subgraph} and develop efficient heuristics to solve the combinatorics.


We evaluate \prjname by training multiple convolutional neural network (CNN) architectures with benchmark datasets such as MNIST~\cite{lecun1998gradient}, CIFAR-10~\cite{krizhevsky2009learning} and ICON-50~\cite{hendrycks2018benchmarking},  and testing on a validation set which consists of original, noisy (i.e., negative of the original images, Gaussian blurred images, images with salt-and-pepper noise) and adversarial images similar to real-world conditions. 
Our results show that the models built by \prjname  did not lose any accuracy against the original clean images, yet delivered better test accuracies against noisy images than the traditional one-hot encoding and prior work, especially when the learning capacity of a model is limited. Our contributions in \prjname  include the following:
  \textbf{a)} novel target encoding scheme based on weighted Hamming distance,  \textbf{b)} effective heuristics for subgraph isomorphism in the target encoding context, and \textbf{c)} comprehensive study on the effects of our target encoding scheme on both noisy and adversarial images.

\section{Related Work}
\label{sec:related}
In this section, we will briefly review various encoding techniques used in deep learning. Please refer to the supplementary material (SM) for an extended review.

\begin{description}
 \item[Multi-class Single-label:] This is the traditional one-hot (1-of-N) encoding scheme that is popularly used in deep learning. One-hot encoding simply represents the target labels numerically without any semantic meaning and is prone to noisy inputs or adversarial attacks~\cite{hendrycks2018benchmarking,szegedy2013intriguing,madry2017towards,goodfellow2014explaining}.
 
 \item[Multi-class Multi-label:] This is the superposition of multiple one-hot encodings for the case where a sample belongs to multiple semantic classes~\cite{hsu2009multi, cisse2013robust}. This contrasts with multi-hot encoding (i.e. \prjname ) where a sample belongs to only one class, but represented in multiple bits.

\item[Label Embedding:] This is a technique to embed target labels with meta information like attributes~\cite{akata2016label} or hierarchies~\cite{tsochantaridis2005large} to capture various structures of the output space at the cost of additional labelling and expert knowledge~\cite{frome2013devise, hsu2009multi}. In some cases, the embedding is jointly learned from input and output data~\cite{amit2007uncovering, weston2010large}.    

\item [Target Encoding:] These are methods that explore alternatives to one-hot encoding~\cite{kim2019multiway,dietterich1994solving, langford2005sensitive, kuncheva2005using, deng2010applying, rodriguez2018beyond}, similar to \prjname. Yet, the key innovations in \prjname  over them are as follows:
\textbf{a)} \prjname  has the same encoding length as one-hot unlike~\cite{kim2019multiway}, and could be used as a drop-in-replacement of any existing one-hot encoding, \textbf{b)} \prjname  strategically optimizes the encoding w.r.t. the given class distribution. Extended review on Target Encoding is in SM.
\end{description}


\section{Data-Similarity Driven Multi-hot Target Encoding}
\label{sec:method}
 
We propose a new target encoding system, \prjname, where multiple output bits are activated. For an $N-$class classification problem   with \prjname, the output layer has $N-$bits, out of which $K$ bits are $1$s as chosen, where $K > 1$. Each output bit has a bounded non-linear activation function and trained such that the binary cross entropy loss between the prediction and target label is minimized. 

The core idea in \prjname is to evaluate the similarity between a given set of classes, then leverage that information to build robust neural networks. To that end, first, the similarity between classes in a given dataset is measured. These similarity weights are used to generate a set of target codes such that the Hamming distances among all classes are maximized and a pair of similar classes are assigned codes with larger Hamming distance based on the extent of similarity. This increases the distance between classes in the feature space as illustrated in Figure~\ref{fig:overview}, thereby reducing the probability of mis-prediction, increasing the generalization of the classifier, and forcing a model to learn the semantics  rather than spatial correlation among pixels. Any chosen neural network could be trained and tested with the generated target encoding without increasing the number of parameters. Overall flow of the encoding generation is in Figure~\ref{fig:flow}, which will be discussed in the following sections.

\subsection{Generating Weights For Target Codes}
\label{sec:weights}

When a deep neural network is trained with back-propagation, it is being optimized to learn features that distinguish classes. When it is able to learn distinct features, these classes are well separated in feature space and have a low probability of mis-prediction (Figure~\ref{fig:overview}(a),(b) and (d)). However, when a network is not able to learn distinguishable features, these classes are not well separated, which may result in poor accuracy (Figure~\ref{fig:overview}(c)). The hypothesis is to identify such visually similar classes and assign target codes that have high Hamming distance between them. By training with such target codes, the network is optimized to learn features to separate similar classes. Having clear decision boundaries between classes help to reduce classification error and better performance with noisy and adversarial images.


The objective of this step is to quantitatively identify similar classes and assign weights to the level of similarity. A confusion matrix ($CM$) is a known method to represent inter-class similarities. $CM$ is essentially a $N \times N$ matrix for a $N-$class dataset with the elements being the confusion metric between a pair of classes as shown in Figure~\ref{fig:flow} (top-left). $CM$ can be obtained by inferencing using a validation set on an already trained target model. We use the method proposed in~\cite{AnonymousTask} to generate $CM$ where the confusion-level between classes of a dataset can be determined by reconstructing data for each class using an autoencoder trained using class $C_i$ and determining the reconstruction error for every other class $C_j$ in the dataset.  
Once $CM$ is obtained for a given dataset, \prjname can convert the confusion matrix into weights (to be used in Section~\ref{sec:code}) in the following method:  the weights are obtained by subtracting the diagonal of $CM$ (self-error values) from $CM$, finding the minimum error of the upper and lower triangles of $CM$, thresholding larger errors to eliminate dissimilar classes, and scaling the non-zero values of $CM$. 



\subsection{Generating Target Codes}
\label{sec:code}

\begin{figure}
  \centering
  \includegraphics[width=0.8\textwidth]{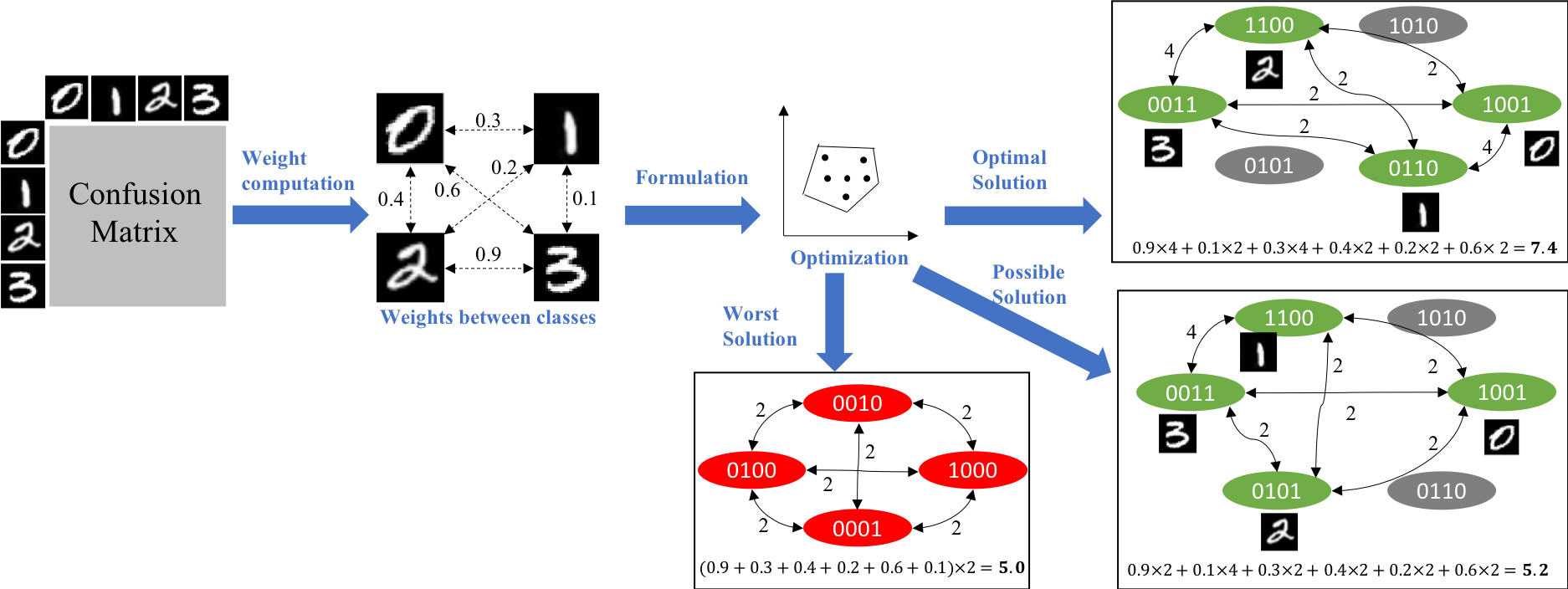}
  \caption{Flow chart of multi-hot encoding generation in \prjname.}
  \label{fig:flow}
\end{figure}

With weights obtained from Section~\ref{sec:weights}, the next step in \prjname is to generate a set of multi-bit target encoding. Our goal is two-fold: \textbf{a}) to generate encoding that maximize Hamming distances among all encoding and \textbf{b}) to assign a pair of encoding that has larger Hamming distance to a pair of more similar classes. In this light, we generate encodings that maximize the total minimum Hamming distance and the Hamming distances between classes based on inter-class similarities.

Inter-class similarities are taken into account in our objective function as weights of Hamming distances over pairs of classes. We put the total minimum Hamming distance and weighted Hamming distances together in our objective function as in Equation~\ref{eq1}.
\begin{equation}
W_{min}H_{min} + \sum\limits_{i=0}^{N-1}\sum\limits_{j=i+1}^{N} W_{ij}H_{ij}
\label{eq1}
\end{equation}
where $W_{min}$ denotes the weight of the minimum Hamming distance $H_{min}$, $N$ denotes the total number of classes, $W_{ij}$ and $H_{ij}$ denote inter-class similarity and the Hamming distance between class $i$ and class $j$, respectively. $N$ and $W_{ij}$ are given constants (i.e., $W_{ij}$ is computed as in Section~\ref{sec:weights}). 
$W_{min}$ should be larger than $W_{ij}$ because maximizing $H_{min}$ drives the overall model performance. We set $W_{min}$ as the number of pairs of classes, $N(N-1)/2$, in our experiments. In one-hot encoding, $H_{min}$ is fixed at $2$, while $H_{min}$ is maximized in \prjname.

Figure~\ref{fig:flow} shows an example of possible outcomes from the optimization step. With one-hot encoding (bottom-center), there exists one trivial solution and the weighted sum of the Hamming distances is the lowest among the solutions illustrated. Figure~\ref{fig:flow} also shows the difference between the optimal and a possible solution in terms of the Hamming distance and the generated encoding: the optimal solution picks a set of encoding and assigns them to the four classes such that more similar classes (i.e., $2$ vs. $3$ with weight $0.9$) are assigned to two codes with the larger Hamming distance.

Optimization in Figure~\ref{fig:flow} can be formulated as an Integer Linear Programming (ILP)  due to the discrete nature of the encoding itself, and solved by a commercial package.
However, it requires a large amount of computation time to find an optimal solution because it explores solution space in an exhaustive branch-and-bound manner. Yet, one useful observation about this combinatorics is that the distribution of Hamming distance values are very discrete and narrow, and has high-density only on a few integer values. Therefore, we expect that there should be a number of possible solutions that have the same objective values. In other words, this maximization problem has a wide and sparse solution space, allowing us to develop an efficient heuristic, \textbf{Narrow-convergence Approach} for \prjname, which is inspired from~\cite{fiduccia1982linear} and finds a solution to Equation~\ref{eq1} significantly faster than ILP optimizations in the following steps:


\begin{itemize}
  \item \textbf{S0:} Equation~(\ref{eq1}) with   $W_{ij}=1$ is optimized as ILP for a given number of classes and bits for a short amount of time until it sufficiently narrows down the solution space. The time limit depends on the number of classes, $N$, and the number of bits, $K$. 
  \
  \item \textbf{S1:} Capture the intermediate result from Step~\textbf{S0} to form a set of initial encodings, $\mathcal{E}$.
 \
 \item \textbf{S2:}  Pick one class randomly without replacement from $\mathcal{E}$.
 
 \item \textbf{S3:}  Choose a small set of alternative encodings for the chosen class, and put them into the candidate pool, $\mathcal{P}$.
 \
 \item \textbf{S4:} Pick the encoding that maximizes Equation~(\ref{eq1}) from $\mathcal{P}$, and assign the encoding to the chosen class. \textbf{S2}-\textbf{S4} is repeated until every class is updated.
 \
 \item \textbf{S5:} Go back to Step~\textbf{S2} until no further improvement in the objective function is found (otherwise exit).
\end{itemize} 


The main difference between ILP optimization and narrow-convergence approach is the strategy for finding solutions. ILP takes a breadth-first search strategy to find an optimal solution, whereas our approach takes a depth-first search strategy. A formal description of Narrow-convergence Approach is available in the supplementary material.

\subsection{Training Method}
\label{sec:train}

\begin{figure}%
    \centering
    \subfloat[\label{fig:train}]{{\includegraphics[width=0.45\textwidth]{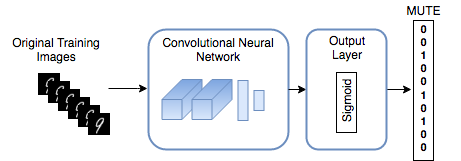} }}%
    \qquad
    \subfloat[\label{fig:test}]{{\includegraphics[width=0.45\textwidth]{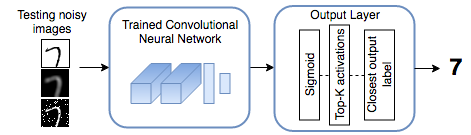} }}%
    \caption{Overview of the (a) training and (b) inferencing methods}%
    \label{fig:example}%
\end{figure}

The \prjname could be used with any CNN model with no changes to the architecture. The traditional softmax classification layer is replaced with a sigmoid layer. The CNN is trained by back-propagating the binary cross entropy loss at each bit. CNN is optimized by SGD with momentum. An overview of training a CNN with \prjname is given in Figure~\ref{fig:train}. In this method, the number of neurons in the CNN and the computational complexity is the same as using one-hot target encoding.

\subsection{Inferencing Method}
\label{sec:test}
As shown in Figure~\ref{fig:test}, an input image is forward propagated through the trained model. The output of the sigmoid layer is thresholded by setting all but the $top-K$ activated bits to $0$, where $K$ is the number of bits used in the \prjname. Then, the Euclidean distance between the thresholded output and each of \prjname encodings are computed. The label corresponding to the closest \prjname code is the classification result.
In the one-hot case, the model has to activate only $1$ bit to deliver a classification. In the proposed method, the model has to activate $K$ output bits to deliver a classification. Even if one or two bits flip due to noise, it may not change the nearest label. Hence, error correcting code \prjname is able to deliver the correct prediction despite noise. Computing Euclidean distance between the set of target encoding may have a potential impact on inference latency. The latency could be minimized by optimizing the Euclidean distance computation using a fast XOR method.

\section{Results \& Discussion}
\label{sec:results}


\subsection{Generation of \prjname}
We compared our Narrow-convergence Approach with ILP optimization using CPLEX~\cite{cplex} in terms of the solution quality and computation time (Table \ref{table:cost}). We broke-down ILP optimization into two approaches: optimization with weights, and optimization without weights followed by shuffling with weights. The experiment was to generate 10bits/4-hot target encodings for $10$ classes. Our experiment environment includes $56$ CPU cores with $120$GB RAM memory. CPLEX ran with parallel threads utilizing $32$ cores in the experiment.

\begin{table}[h]
	\centering
	\small
	\begin{tabular}{|c|c|c|c|c|} 
		\hline
		& Optimization approach & Objective value & Min. Hamming distance & Computation time \\
		\hline
		\hline
		\multirow{3}{*}{\begin{tabular}[x]{@{}c@{}}10-bit\\4-hot\end{tabular}} & Narrow-convergence & 450.21 & 4 & 90 sec\\
		\cline{2-5}
		& CPLEX (w/ weights) & 450.31 & 4 &  > 11 hrs \\  
		\cline{2-5}
		& CPLEX (w/o weights) & 447.75 & 4 & > 50 hrs \\
		\hline
	\end{tabular}
	\caption{Ours reduced run-time by more than $99.75\%$ compared to conventional ILP.}
	\label{table:cost}
\end{table}


For the 4-hot/10bits case with weights derived from MNIST dataset, narrow-convergence approach reduced the time cost by more than $99.75\%$, while it performs as good as ILP optimization. ILP with weighted objective function takes more than $11$ hours to reach an objective value of $450.31$. We were not able to run more than $11$ hours due to memory limitations. Solving unweighted objective function for $50$ hours followed by shuffling with weights reaches an objective value of $447.75$.

\subsection{Performance of \prjname}
\label{sec:mute_exp}
\textbf{Datasets.}~We used the MNIST~\cite{lecun1998gradient}, CIFAR-10~\cite{krizhevsky2009learning} and ICON-50~\cite{hendrycks2018benchmarking} datasets for experiments. MNIST is a dataset of handwritten digits in black and white containing $60,000$ training images and $10,000$ testing images. CIFAR-10 is a of dataset of $32 \times 32$ color images belonging to $10$ classes representing airplanes, cars, birds, cats, deer, dogs, frogs, horses, ships, and trucks. It has $50,000$ training and $10,000$ testing images equally distributed among its classes. ICON-50 is a set of $10,000$ color icons of size $32 \times 32$ belonging to $50$ classes such as airplane, ball, drink, feline, etc., and collected from various companies such as, Apple, Microsoft, Google, Facebook, etc. The icons from different companies exhibit different styles and versions. This is a challenging dataset because there is a class imbalance between icons collected from various companies (in other words different styles). 

\textbf{CNN Architectures.}~We utilized a wide range of CNN architectures in our experiments. MNIST dataset was trained and tested with two CNNs: LeNet~\cite{lecun1998gradient} and ConvNet. ConvNet has $2$ convolutional layers with $5 \times 5$ kernel size, followed by a fully connected layer with $50$ neurons and a final sigmoid layer. CIFAR-10 and ICON-50 datasets are trained and tested with AlexNet~\cite{krizhevsky2012imagenet}, DenseNet~\cite{huang2017densely}, ResNet~\cite{he2016deep}, and ResNeXt~\cite{xie2017aggregated} architectures. DenseNet has depth of $40$ and growth rate of $24$. ResNet has depth of $20$. ResNeXt has depth of $29$ and cardinality of $8$.

\begin{figure}[t]
	\centering
	\includegraphics[width=\textwidth]{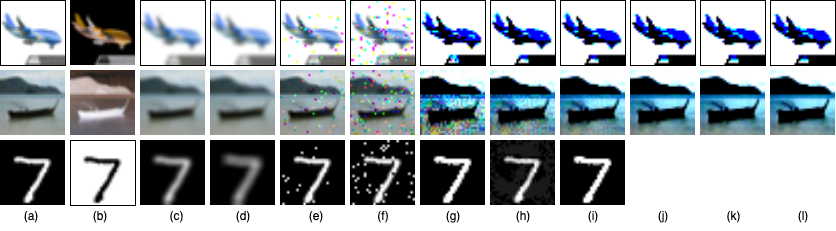}
	\caption{Images from the various test sets. The images in the first row is from the ICON-50~\cite{hendrycks2018benchmarking} dataset, second row is from the CIFAR-10~\cite{krizhevsky2009learning} dataset and the third row is from the MNIST~\cite{lecun1998gradient} dataset. (a) Original images. (b) Negative~\cite{hosseini2017limitation} of the original images. (b - d) Original images blurred with Gaussian kernels of $\sigma = 1$ and $\sigma = 2$ respectively. (e - f) Salt \& pepper noise added to the original images at $2\%$ and $5\%$ respectively. (g - l) Adversarial images with FGSM~\cite{goodfellow2014explaining} noise with $\epsilon = 0.2, 0.1, 0.05, 0.01, 0.005$ and $0.001$ respectively.}
	\label{fig:test_data}
\end{figure}

\textbf{Experiments.}~We used the standard train/test split for MNIST and CIFAR-10 datasets. For ICON-50, $80\%$ of images in each class were randomly assigned to the training set and rest were allocated to the test set. Various CNN architectures were trained with original images in the training dataset for $200$ epochs. The trained models were tested with original images in the test set. Additionally, to evaluate the robustness of the trained models, we created and tested noisy versions of the original test sets as shown in Figure~\ref{fig:test_data}. Negative images~\cite{hosseini2017limitation} have the same spatial structure as the original images but are in a diagonally opposite color space. These images are useful to evaluate if the trained models have learned the semantic structure of the dataset or merely memorized the spatial pixel correlations in the training images. Typical noisy images such as Gaussian blurred images with $\sigma = 1$ and $2$ and Salt \& Pepper noise at $2\%$ and $5\%$ were created. To test robustness against adversarial attacks, we created adversarial images with Fast Gradient Sign Method (FGSM)~\cite{goodfellow2014explaining} with $\epsilon = 0.05, 0.1$ and $0.2$. This creates a comprehensive test set that evaluates models for their generalization and robustness capability.

\begin{figure}[t]
	\hspace{-0.45in}
 	\includegraphics[width=6.2in]{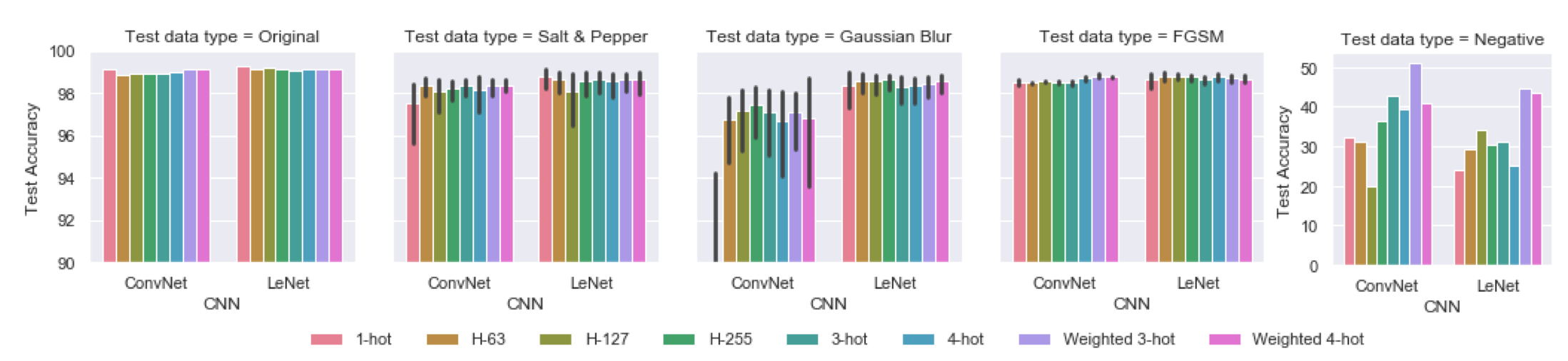}
	\caption{\prjname with LeNet and ConvNet architectures trained on MNIST data improves the one-hot average test accuracy by $2.8\%$ and $7.1\%$ respectively. Whereas Hadamard target encoding improves one-hot performance only by $1.7\%$ and $3.5\%$.}
	\label{fig:mnist}
\end{figure}

\begin{figure}[t]
	\hspace{-0.45in}
	\includegraphics[width=6.2in]{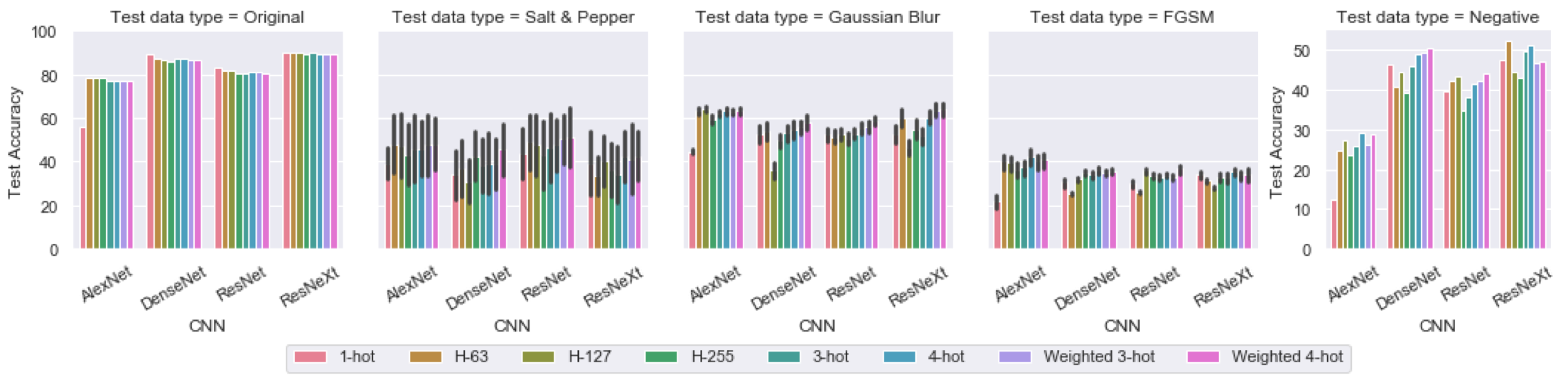}
	\caption{\prjname encoding with AlexNet, DenseNet, ResNet and ResNeXt architectures trained on CIFAR-10 data improves the one-hot average test accuracy by $41.5\%$ , $4.2\%$, $3.6\%$ and $3.8\%$ respectively. Whereas Hadamard target encoding improves one-hot performance only when used with AlexNet and ResNet architecture. When used with DenseNet and ResNeXt, the average test accuracy is worse than using one-hot encoding. }
	\label{fig:cifar10}
\end{figure}

\begin{figure}[]
	\hspace{-0.45in}
	\includegraphics[width=6.2in]{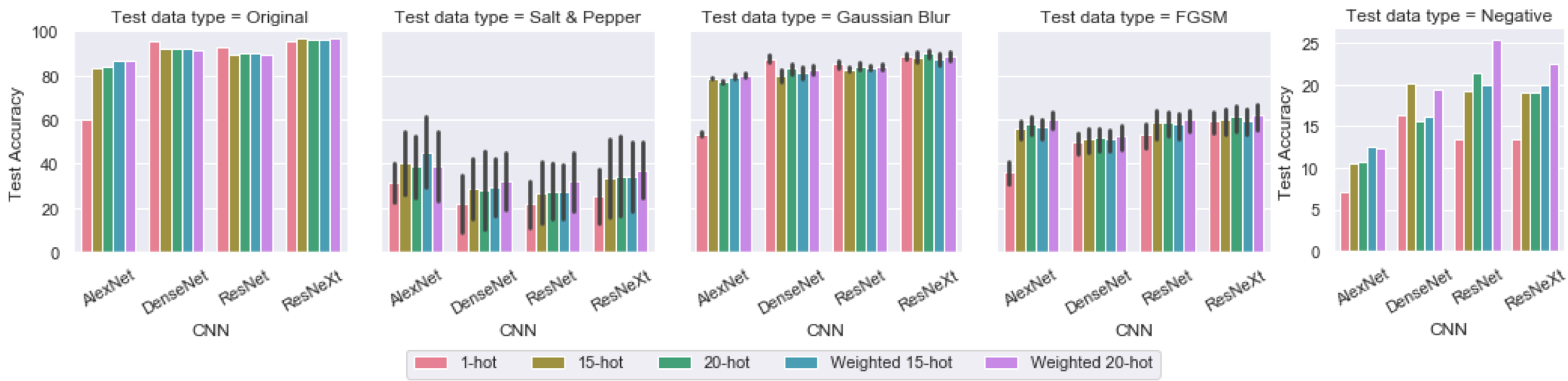}
	\caption{AlexNet, DenseNet, ResNet, and ResNeXt architectures trained with various \prjname on ICON-50 data and tested with the original and noisy testsets improved average test accuracy over one-hot by $42.29\%$, $0.37\%$,  $4.70\%$, and  8.12$\%$ respectively.}
	\label{fig:icon50_all}
\end{figure}

\begin{figure}[h]
	\centering
	\vspace{-0.15in}
	\includegraphics[width=0.8\textwidth]{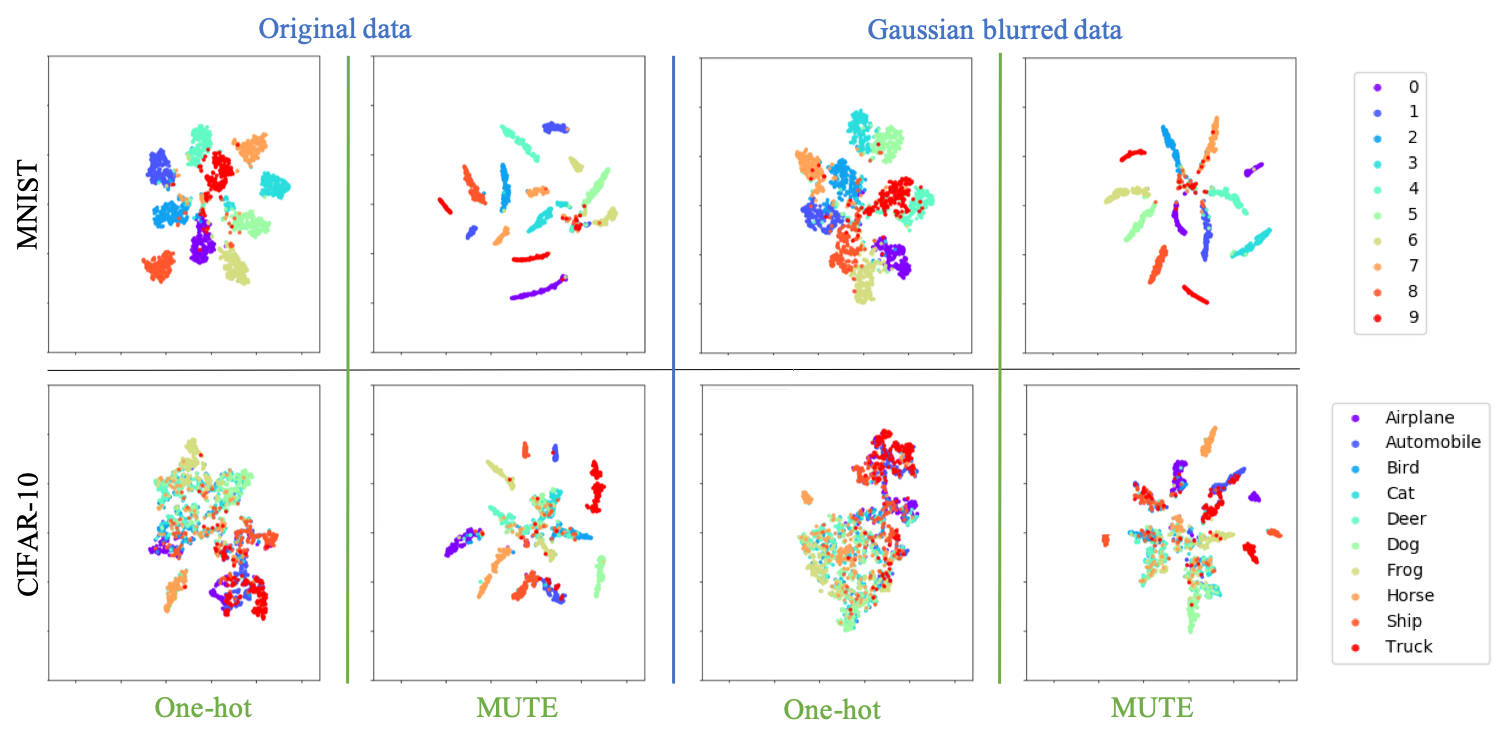}
	\caption{Features learned with \prjname are well-separated and condensed in multi-dimensional space such that model maintains separation even when tested with noisy data. In comparison, features learned by one-hot encodings lack isolation in feature space.}
	\label{fig:tsne_comp}
\end{figure}

In our experiments, we compared against the conventional one-hot encoding and Hadamard target encoding methods. In addition to comparing with Hadamard encoding results in~\cite{yang2015deep}, we substituted the proposed \prjname with Hadamard encodings~\cite{yang2015deep} and trained and tested on the chosen datasets. Hadamard results are shown as H-63, H-127 and H-255 in the Figures~\ref{fig:mnist} and ~\ref{fig:cifar10}. We also conducted experiments with weighted and unweighted \prjname for different number of hot bits. The results of our experiments are shown in Figures~\ref{fig:mnist},~\ref{fig:cifar10},~\ref{fig:icon50_all},~\ref{fig:tsne_comp} and Table~\ref{tab:icon50_subtype}.

We trained models with various target encodings with the same set of hyper-parameters to eliminate the impact of hyper-parameter choice on the model performance. We did not augment MNIST, however augmented CIFAR-10 and ICON-50 data using randomized affine transformations, color jitter and cropping in random order. The augmentation parameters were chosen by trial and error. The batch size was $128$. CNNs were optimized using SGD with $0.9$ momentum, $0.0001$ weight decay, and $0.1$ initial learning rate. All experiments were conducted on an Intel(R) machine with Xeon(R) CPU E5-2680 v4 at $2.40$ GHz frequency. It has $504$ GB RAM and $56$ CPU cores. The machine has $4$ Tesla P100-PCIE GPUs. Models were trained using only $1$ GPU at a time. Code was written using PyTorch V1~\cite{paszke2017automatic} implementation.

Figure~\ref{fig:mnist} shows the test accuracy of LeNet and ConvNet architectures with different target encodings trained on original images in the MNIST training dataset and tested on original, noisy and adversarial versions of the MNIST test dataset. The barplots illustrate the central tendency for different test datasets and uncertainty (error bars) for test images impacted by varying amounts of Gaussian blur, Salt \& Pepper noise and FGSM. The proposed \prjname method has better average test performance than one-hot encoding or Hadamard target encoding (H-63, H-127, and H-255). The best test accuracy on original images reported by~\cite{yang2015deep} using Hadamard Codes on MNIST is $85.47\%$ using direct classification with H-255. The authors~\cite{yang2015deep} use a CNN architecture with $3$  convolutional layers, $1$ locally connected layer and $2$ fully-connected layers. The proposed \prjname method improves this result by $13.61$ percentage points on average.

CNN architectures with various \prjname were trained on original images in the CIFAR-10 training dataset and tested on original and noisy versions of the CIFAR-10 test dataset as shown in Figure~\ref{fig:cifar10}. The proposed \prjname method with AlexNet, DenseNet, ResNet and ResNeXt architectures improved average test accuracy over one-hot encoding. However, choice of architecture seems to be important when using Hadamard encodings, as performance dropped when using DenseNet and ResNeXt architectures. \prjname is particularly useful when a less robust architecture such as AlexNet~\cite{hendrycks2018benchmarking} is used.

\begin{table}[]
	\centering
	\resizebox{0.75\textwidth}{!}{%
		\begin{tabular}{|c|c|c|c|c|c|}
			\hline
			\multirow{2}{*}{CNN Arch.} & \multicolumn{5}{c|}{Target Encoding}                        \\ \cline{2-6} 
			& 1-hot & 15-hot & 20-hot & Weighted 15-hot & Weighted 20-hot \\ \hline
			\hline
			AlexNet                    & 21.85 & 33.78  & \textbf{34.45}  & 32.72           & 32.83           \\ \hline
			DenseNet                   & 31.02 & 34.07  & 33.72  & 31.77           & \textbf{34.95}           \\ \hline
			ResNet                     & 29.87 & 32.31  & 31.56  & 32.21           & \textbf{33.31}           \\ \hline
			ResNeXt                    & 34.68 & \textbf{34.86}  & 34.66  & 34.51           & 34.76           \\ \hline
		\end{tabular}%
	}
\caption{Average test accuracy on noisy data of the ICON-50 sub-type robustness experiment in~\cite{hendrycks2018benchmarking}.}
\label{tab:icon50_subtype}
\end{table}

We trained AlexNet, DenseNet, ResNet, and ResNeXt architectures on an ICON-50 training set that contained images from all $50$ classes and tested on original and noisy images in the test set (Figure~\ref{fig:icon50_all}). \prjname increased one-hot encoding performance $0.3\%$ to $42.29\%$ with AlexNet, the least robust CNN~\cite{hendrycks2018benchmarking}, benefiting the most from \prjname. In another experiment using Icon-50 dataset, we trained the CNN architectures on all original images belonging to the sub-type robustness experiment in~\cite{hendrycks2018benchmarking}. Then tested on the noisy versions (negative, Gaussian blurred, and Salt \& Pepper noise) of the held-out sub-types. Table~\ref{tab:icon50_subtype} lists the average test accuracy on the noisy test sets. \prjname consistently obtains higher average test accuracy for any number of hot bits or weighted configuration.
We used the TSNE~\cite{maaten2008visualizing} algorithm to visualize ConvNet and AlexNet models trained with one-hot and \prjname on MNIST and CIFAR-10 datasets, respectively. Figure~\ref{fig:tsne_comp} shows that models trained with one-hot encodings fail to learn distinguishable features that give rise to good decision boundaries. Thus, noisy data is easily mis-predicted. On the other-hand, condensed features learned by \prjname enable clear decision boundaries even when faced with noisy data.

\section{Conclusion}
\label{sec:conc}

We propose a novel target encoding methodology that is cognizant of inter-class similarities to train neural networks. We proposed Narrow-convergence Approach, a time-effective method to generate \prjname compared to ILP optimization. We show that \prjname can be used to train neural networks that are robust to different noisy and adversarial images while maintaining comparable accuracy on original images, yet has the same computation complexity as using one-hot target encoding.



{
\small
\bibliographystyle{ieee}
\bibliography{code}
}


\end{document}